\documentclass{article}

\PassOptionsToPackage{numbers,sort&compress}{natbib}
\usepackage[sglblindworkshop]{neurips_2025}
\workshoptitle{AICC - AI for Climate and Conservation}

\usepackage[utf8]{inputenc}
\usepackage[T1]{fontenc}
\usepackage{hyperref}
\usepackage{url}
\usepackage{booktabs}
\usepackage{amsfonts}
\usepackage{nicefrac}
\usepackage{microtype}
\usepackage{xcolor}
\usepackage{amsmath}
\usepackage{amssymb}
\usepackage{graphicx}
\usepackage{tikz}
\usetikzlibrary{shapes,arrows,positioning,fit,backgrounds}

\title{Hierarchical Re-Classification: 
Combining Animal Classification Models with Vision Transformers}

\author{%
  Hugo Markoff$^{*}$ \\
  CTO Animal Detect \\
  Animal Detect \\
  Aalborg, Denmark \\
  \texttt{hugo@animaldetect.com} \\
  \And
  Jevgenijs Galaktionovs \\
  CEO Animal Detect \\
  Animal Detect \\
  Aalborg, Denmark \\
  \texttt{eugene@animaldetect.com} \\
}

\usepackage{etoolbox}

\AtBeginDocument{\nolinenumbers}

\begin{document}

\maketitle

\noindent $^{*}$Corresponding author

\begin{abstract}
State-of-the-art animal classification models like SpeciesNet provide predictions 
across thousands of species but use conservative rollup strategies, resulting in 
many animals labeled at class or kingdom level rather than species. 
This work presents a hierarchical re-classification system developed for the 
Animal Detect platform~\cite{animaldetect}, combining SpeciesNet's 
EfficientNetV2-M predictions with CLIP embeddings and metric learning to refine 
high-level taxonomic labels toward species-level identification.

We evaluate our five-stage pipeline (high-confidence acceptance, bird override, 
centroid building, triplet-loss metric learning, and adaptive cosine-distance scoring) 
on a segment of the LILA BC Desert Lion Conservation dataset (4,018 images, 15,031 detections). 
After recovering 761 bird detections from "blank" and "animal" labels, we re-classify 456 detections labeled animal, 
mammal, or blank with 96.5\% accuracy, achieving species-level identification for 
64.9\%. 
\end{abstract}

\section{Introduction}

State-of-art wildlife classifiers can achieve high accuracy on a wide range of
species but struggle with unknown species or similar species \citep{green2020automated}.
Google's SpeciesNet model \citep{cameratrapai,speciesnet_kaggle} uses an EfficientNetV2-M backbone \citep{tan2021efficientnetv2}
and an ensemble strategy with optional geofencing and conservative taxonomic rollup, which often results in
animals labeled at class or kingdom level rather than species.

Manual review of these high-level taxonomic images creates annotation bottlenecks, 
particularly problematic when rapid assessment is critical \cite{glover2019camera}.
To address this challenge, we developed a hierarchical re-classification system 
for Animal Detect~\cite{animaldetect}, a camera trap analysis platform designed 
to streamline wildlife monitoring workflows.

This work addresses: \emph{Can we automatically rolldown high taxonomic predictions closer to species-level by leveraging visual similarity in learned embedding 
spaces?} 
We hypothesize that even when a classifier lacks confidence for species-level 
identification, images might cluster near \emph{accepted} high-confidence predictions 
of the same species in embedding space.

We present a hierarchical re-classification system that successfully re-classifies 
high-level taxonomic predictions and achieves species-level identification for 
many images initially labeled with generic categories like "animal" or "mammal". 
The system is deployed in production in Animal Detect~\cite{animaldetect} with 
interactive sorting and suggestion features that enhance human-in-the-loop 
annotation workflows for conservation researchers worldwide.

\section{Background}

\textbf{SpeciesNet} provides classification across 
thousands of species using EfficientNetV2-M \cite{tan2021efficientnetv2}, 
but deliberately returns conservative predictions (e.g., "animal", "mammal") 
when species-level confidence is insufficient, prioritizing 
precision over recall.

\textbf{Our approach.} We leverage high-confidence predictions as anchors for 
metric learning, using visual similarity through 
taxonomic-level-aware scoring respecting hierarchical relationships. 
Triplet loss \cite{schroff2015facenet} learns embeddings where same-species 
images cluster tightly while different species separate.

\section{Method}

\subsection{Pipeline Architecture}

\begin{figure}[h]
\centering
\small
\begin{tikzpicture}[
    node distance=0.35cm and 0.5cm,
    box/.style={rectangle, draw, thick, minimum width=1.4cm, minimum height=0.6cm, align=center, font=\tiny},
    process/.style={box, fill=blue!10},
    output/.style={box, fill=green!10},
    input/.style={box, fill=orange!10},
    reclass/.style={box, fill=purple!10, minimum width=1.6cm, minimum height=1.0cm},
    arrow/.style={->, >=stealth, thick},
]

\node[input] (image) {Image};
\node[process, right=0.4cm of image] (detector) {Detector};
\node[process, right=0.4cm of detector] (hitl) {HITL};
\node[process, right=0.4cm of hitl] (crop) {Crop};

\node[process, above right=0.25cm and 0.4cm of crop] (speciesnet) {SpeciesNet};
\node[output, right=0.35cm of speciesnet] (pred) {Ensemble};

\node[process, below right=0.25cm and 0.4cm of crop] (clip) {CLIP};
\node[output, right=0.35cm of clip] (embed) {Embed};

\node[reclass, right=0.25cm of pred, yshift=-0.7cm] (reclass) {\textbf{Re-class}\\{\scriptsize 5-stage}};

\draw[arrow] (image) -- (detector);
\draw[arrow] (detector) -- (hitl);
\draw[arrow] (hitl) -- (crop);
\draw[arrow] (crop.north) |- (speciesnet.west);
\draw[arrow] (crop.south) |- (clip.west);
\draw[arrow] (speciesnet) -- (pred);
\draw[arrow] (clip) -- (embed);
\draw[arrow] (pred) -| (reclass.north);
\draw[arrow] (embed) -| (reclass.south);

\end{tikzpicture}
\caption{Pipeline: Detection → HITL → dual-path (SpeciesNet + CLIP) → 
5-stage re-classification}
\label{fig:pipeline}
\end{figure}
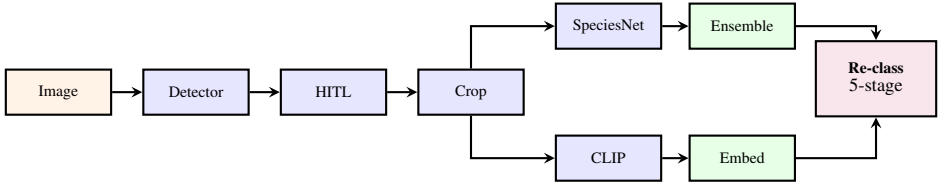

Figure~\ref{fig:pipeline} illustrates the Animal Detect pipeline until and including 
the classification stage. 
MegaDetector v5a \citep{megadetector_github} provides binary animal detection
with human-in-the-loop review. 
Each crop processes through: (1) SpeciesNet producing species predictions 
with ensemble logic, (2) CLIP ViT-L/14 \citep{radford2021learning,dosovitskiy2021an} extracting 768-dimensional embeddings.
Five-stage re-classification then upgrades high-level taxonomic predictions.

\subsection{SpeciesNet Ensemble Strategy}

SpeciesNet combines MegaDetector (object detector) with EfficientNetV2-M 
(species classifier) through multi-step ensemble logic optimized to minimize 
human review burden~\cite{cameratrapai}. 
The workflow prioritizes: (1) detection-based human/vehicle override when 
detector confidence exceeds threshold, (2) blank identification combining 
classifier and detector signals, (3) geofencing to filter geographically 
impossible species, (4) \textit{taxonomic rollup} propagating predictions to 
genus, family, order, class, or kingdom when species-level confidence 
insufficient, (5) detector-based ``animal'' fallback, (6) ``unknown'' last 
resort. 
This conservative strategy produces many high-level taxonomic labels,
precisely what our re-classification system targets to refine toward 
species-level identification.

\subsection{Five-Stage Re-Classification}

\textbf{Step 1: High-confidence acceptance.} Images with 
\( s_{\text{SpeciesNet}} \geq 0.8 \) immediately accepted as anchors.

\textbf{Step 2: Bird override.} Rolldown to class, from kingdom labeled when \(\geq 3\) 
bird species in top-5 pre-ensemble predictions and summed confidence 
\(\geq 0.3\).

\textbf{Step 3: Centroid building.} For each accepted label \(c\) with 
\(\geq 5\) samples, compute centroid 
\(\boldsymbol{\mu}_c = \frac{1}{|\mathcal{C}_c|} \sum_{\mathbf{e} \in 
\mathcal{C}_c} \mathbf{e}\). 
Accept additional crops where nearest centroid matches predicted species and 
cosine distance \(\leq\) 95th percentile of in-cluster distances.

\textbf{Step 4: Metric learning.} Train neural network 
\(\mathbf{f}_\theta : \mathbb{R}^{768} \rightarrow \mathbb{R}^{256}\) using 
triplet loss with margin \(\alpha = 1.0\):
\begin{equation}
\mathcal{L}_{\text{triplet}} = \sum_{i=1}^{N} \Bigl[ 
\|\mathbf{f}_\theta(\mathbf{e}_i^a) - \mathbf{f}_\theta(\mathbf{e}_i^p)\|_2^2 
- \|\mathbf{f}_\theta(\mathbf{e}_i^a) - \mathbf{f}_\theta(\mathbf{e}_i^n)\|_2^2 
+ \alpha \Bigr]_+
\end{equation}
Embeddings L2-normalized: 
\(\mathbf{e}' = \mathbf{f}_\theta(\mathbf{e}) / \|\mathbf{f}_\theta(\mathbf{e})\|_2\).

\textbf{Step 5: Adaptive scoring.} Re-classify using taxonomic-level-aware 
cosine distance with cluster-tightness weighting:
\begin{equation}
\text{score}_c(\mathbf{e}) = \frac{1 - \hat{\boldsymbol{\mu}}_c^\top \mathbf{e}'}
{\bar{d}_c \cdot w_c}, 
\quad w_c = 1 + \frac{\bar{d}_{\max} - \bar{d}_c}{\bar{d}_{\max}}
\end{equation}
where \(\hat{\boldsymbol{\mu}}_c = \boldsymbol{\mu}_c' / \|\boldsymbol{\mu}_c'\|_2\) (renormalized centroid), 
\(\bar{d}_c\) is average intra-cluster cosine distance, and \(w_c\) weights by cluster tightness. 
Images with \(\text{score}_c < \tau = 0.85\) (empirically chosen to maximize F\(_1\) with precision \(\geq\) 95\%) and taxonomic hierarchy match get re-classified.

\section{Experiments}

\textbf{Dataset.} We evaluate on a randomly sampled subset from the LILA BC 
Desert Lion Conservation dataset: 4,018 images yielding 15,031 detections 
across African savanna species. 
Initial SpeciesNet predictions: 2,862 detections labeled "animal", 891 "blank", 
288 "mammal" (of 15,031 total detections). 
After processing, we manually validated all re-classified images against 
ground truth. 

\section{Results}

\subsection{Overall Performance}

Table~\ref{tab:combined_results} shows re-classification performance. 
Of 456 detections re-classified (from images initially labeled animal (2,862), mammal (288), or blank (891)), \textbf{440 were correct (96.5\%)}, with 
\textbf{296 (64.9\%) reaching species-level}, upgrading from generic labels. 
Step 2 (bird override) recovered 761 bird detections from "blank" and "animal" pools with 96.8\% accuracy.

\begin{table}[h]
\centering
\caption{Re-classification performance by original label}
\label{tab:combined_results}
\small
\begin{tabular}{@{}lrrrrr@{}}
\toprule
\textbf{Original Label} & \textbf{Re-classified} & \textbf{To Species} & \textbf{Correct} & \textbf{Incorrect} & \textbf{Accuracy} \\
\midrule
Animal & 341 & 221 & 334 & 7 & 97.9\% \\
Mammal & 81 & 53 & 79 & 2 & 97.5\% \\
Blank & 34 & 22 & 27 & 7 & 79.4\% \\
\midrule
\textbf{Total} & \textbf{456} & \textbf{296} & \textbf{440} & \textbf{16} & \textbf{96.5\%} \\
\bottomrule
\end{tabular}
\end{table}

The majority of re-classifications originated from the generic ``animal'' label 
(341 images, 74.8\% of total re-classifications), with 221 (64.8\%) successfully 
upgraded to species-level identification achieving 97.9\% accuracy. 
Re-classifications from ``mammal'' labels achieved similarly high accuracy (97.5\%), 
reflecting the additional taxonomic constraint that narrows the search space. 
Images initially labeled ``blank'' proved most challenging (79.4\% accuracy), 
with 7 of 34 re-classifications incorrect/uncertain, several of the re-classified blank
images were so unclear that even we struggled to classify them confidently and ultimately left them
as false positives.

\subsection{Conservative Rollup Strategy}

Our re-classification pipeline employs a conservative approach, similar to SpeciesNet's 
rollup philosophy. 
During testing, \textbf{84 detections were rolled up to generic ``animal''} 
rather than assigned species-level labels because they failed to meet confidence 
thresholds across all five pipeline stages.

These 84 cases represent potential false negatives detections that might have 
correct species labels but lack sufficient evidence for our pipeline to accept them. 
The conservative rollup strategy prioritizes precision over recall: by accepting 
coarser taxonomic labels for uncertain cases, we maintain high accuracy (96.5\%) 
on the 456 detections we \emph{did} re-classify, while avoiding potentially 
incorrect species assignments. 

\subsection{User Interface Integration}

The system deploys in Animal Detect with two features: 
(1) \textbf{default sorting} by Euclidean distance in metric-learned space 
showing visual similarity across taxonomic levels, helping manually separate 
morphologically similar species, 
(2) \textbf{interactive suggestions} showing best-scoring species when 
right-clicking images not automatically re-classified, providing probabilistic 
guidance without forcing automated decisions. 
These integrate re-classification into human-in-the-loop workflows.

\section{Discussion}

\textbf{Why this works.} SpeciesNet's high-confidence predictions provide 
reliable anchors. 
CLIP embeddings trained on natural images capture morphological similarity 
enabling distance-based refinement. 
Five-stage pipeline progressively refines predictions with increasing 
selectivity.

\textbf{Current limitations.} The cosine-distance scoring is heuristic; 
investigating Mahalanobis distance, learned scoring functions, or Bayesian confidence 
intervals could improve rigor and interpretability. 
System mixes taxonomic levels in single embedding space; separate clustering 
at each level might reduce variance.

\section{Planned Work}

We will systematically address limitations and explore improvements:

\textbf{Architecture comparison.} Benchmark more capable ViT models across 
multiple taxonomic groups to optimize accuracy-speed tradeoffs for production 
deployment.

\textbf{Adaptive threshold optimization.} Investigate class-specific thresholds 
and precision-recall curves across different cluster properties (size, 
tightness, taxonomic level) to maximize annotations upgraded while maintaining conservation-critical accuracy 
requirements.

\textbf{Hierarchical taxonomic processing.} Develop separate clustering 
pipelines at each taxonomic level (family → genus → species) rather than 
mixing levels in single embedding space. 
This leverages biological structure to reduce variance and could improve 
fine-grained discrimination while maintaining interpretable hierarchical 
relationships.

\textbf{Principled scoring functions.} Replace heuristic adaptive scoring 
with theoretically grounded alternatives: (1) Mahalanobis distance accounting 
for cluster covariance structure, (2) learned scoring functions trained 
end-to-end with classification objective, (3) Bayesian confidence intervals 
providing probabilistic guarantees, (4) ensemble scoring combining multiple 
distance metrics.

\textbf{Alternative metric learning.} Explore methods beyond triplet loss 
function for improved margin-based separation, prototypical networks for 
few-shot scenarios, contrastive learning with hard negative mining to improve 
embedding discriminability for morphologically similar species that currently 
confuse the system.

\textbf{Handling conservative rollups.} Investigate the detections 
rolled up to ``animal'' due to insufficient confidence across all pipeline stages. 
These cases warrant dual investigation: (1) determining what proportion 
represents genuinely difficult cases versus SpeciesNet misclassifications in 
the original predictions, and (2) developing methods to safely recover 
true positives while filtering actual errors.

\textbf{Open-source benchmarking.} Release standardized evaluation datasets, 
code, and baseline results to enable reproducible comparison with alternative 
re-classification approaches. 
Establish comprehensive benchmarks across diverse animal data and taxonomic
groups to validate generalization beyond African savanna species. 
Create open-access annotated datasets specifically for hierarchical 
re-classification tasks, including challenging cases with ground-truth labels 
at multiple taxonomic levels

\section{Conclusion}

This work demonstrates hierarchical re-classification combining foundation models, 
Vision Transformers and with metric learning achieve 96.5\% accuracy upgrading
high-level taxonomic wildlife predictions, with 64.9\% reaching species-level identification.
By reducing manual annotation burden and maintaining high precision. 
The five-stage pipeline provides a reusable framework for refining foundation 
model predictions across domains, with planned extensions to more capable 
architectures, adaptive thresholding, and hierarchical taxonomic processing 
to further improve production-system reliability.

\bibliographystyle{unsrtnat}
\bibliography{references}

\end{document}